\documentclass{article}

\usepackage{colacl}
\usepackage{epsf}

\newcommand{\nprule}[1]{$\langle${#1}$\rangle$}
\newcommand{\eg}[1]{\textit{#1}}
\newcommand{\br}[1]{$[${#1}$]$}
\newcommand{\brnp}[1]{$[_{\mathrm{NP}}$~{#1}$]$}

\title{Error-Driven Pruning of Treebank Grammars \\ for Base Noun Phrase Identification}
\author{Claire Cardie and David Pierce \\
Department of Computer Science \\ Cornell University \\
Ithaca, NY~~14853\\ cardie, pierce@cs.cornell.edu}

\begin{document}

\maketitle

\begin{abstract}
Finding simple, non-recursive, base noun phrases is an important
subtask for many natural language processing applications.  While
previous empirical methods for base NP identification have been rather
complex, this paper instead proposes a very simple algorithm that is
tailored to the relative simplicity of the task. In particular, we
present a corpus-based approach for finding base NPs by matching
part-of-speech tag sequences.  The training phase of the algorithm is
based on two successful techniques: first the base NP grammar is read
from a ``treebank'' corpus; then the grammar is improved by selecting
rules with high ``benefit'' scores.  Using this simple algorithm with
a naive heuristic for matching rules, we achieve surprising accuracy
in an evaluation on the Penn Treebank Wall Street Journal.
\end{abstract}

\noindent
\begin{picture}(0,0)
\put(0,370){In \emph{Proceedings of COLING-ACL'98}, pages 218--224.}
\end{picture}

\section{Introduction}
\label{introduction}

Finding base noun phrases is a sensible first step for many natural
language processing (NLP) tasks: Accurate identification of base noun
phrases is arguably the most critical component of any partial parser;
in addition, information retrieval systems rely on base noun phrases
as the main source of multi-word indexing terms; furthermore, the
psycholinguistic studies of Gee and Grosjean \shortcite{gee:chunks}
indicate that text chunks like base noun phrases play an important
role in human language processing.  In this work we define \emph{base
NPs} to be simple, nonrecursive noun phrases --- noun phrases that do
not contain other noun phrase descendants.  The bracketed portions of
Figure \ref{fig:examples}, for example, show the base NPs in one
sentence from the Penn Treebank Wall Street Journal (WSJ) corpus
\cite{marcus:treebank}.  Thus, the string \eg{the sunny confines of
resort towns like Boca Raton and Hot Springs} is too complex to be a
base NP; instead, it contains four simpler noun phrases, each of which
is considered a base NP: \eg{the sunny confines}, \eg{resort towns},
\eg{Boca Raton}, and \eg{Hot Springs}.

\begin{figure}
\begin{center}
When \br{it} is \br{time} for \br{their biannual powwow} , \\ \br{the
nation} 's \br{manufacturing titans} typically \\ jet off to \br{the
sunny confines} of \br{resort towns} \\ like \br{Boca Raton} and
\br{Hot Springs}.
\end{center}
\caption{Base NP Examples}
\label{fig:examples}
\end{figure}

Previous empirical research has addressed the problem of base NP
identification.
Several algorithms identify ``terminological phrases'' --- certain
base noun phrases with initial determiners and modifiers removed:
Justeson \& Katz \shortcite{justeson:terms:nle:1995} look for repeated
phrases; Bourigault \shortcite{bourigault:nps} uses a handcrafted noun
phrase grammar in conjunction with heuristics for finding maximal
length noun phrases; Voutilainen's NPTool \shortcite{voutilainen:nps}
uses a handcrafted lexicon and constraint grammar to find
terminological noun phrases that include phrase-final prepositional
phrases.  Church's PARTS program \shortcite{church:nps}, on the other
hand, uses a probabilistic model automatically trained on the Brown
corpus to locate core noun phrases as well as to assign parts of
speech. More recently, Ramshaw \& Marcus
\shortcite{ramshaw:chunking-long} apply transformation-based learning
\cite{brill:rules-cl} to the problem. Unfortunately, it is difficult
to directly compare approaches.  Each method uses a slightly different
definition of base NP. Each is evaluated on a different corpus.  Most
approaches have been evaluated by hand on a small test set rather than
by automatic comparison to a large test corpus annotated by an
impartial third party.  A notable exception is the Ramshaw \& Marcus
work, which evaluates their transformation-based learning approach on
a base NP corpus derived from the Penn Treebank WSJ, and achieves
precision and recall levels of approximately 93\%.

This paper presents a new algorithm for identifying base NPs in an
arbitrary text.  Like some of the earlier work on base NP
identification, ours is a trainable, corpus-based algorithm.  In
contrast to other corpus-based approaches, however, we hypothesized
that the relatively simple nature of base NPs would permit their
accurate identification using correspondingly simple methods.  Assume,
for example, that we use the annotated text of
Figure~\ref{fig:examples} as our training corpus.  To identify base
NPs in an unseen text, we could simply search for all occurrences of
the base NPs seen during training --- \eg{it}, \eg{time}, \eg{their
biannual powwow}, \ldots, \eg{Hot Springs} --- and mark them as base
NPs in the new text.  However, this method would certainly suffer from
data sparseness.  Instead, we use a similar approach, but back off
from lexical items to parts of speech: we identify as a base NP any
string having the same part-of-speech tag sequence as a base NP from
the training corpus.  The training phase of the algorithm employs two
previously successful techniques: like Charniak's
\shortcite{charniak:treebank} statistical parser, our initial base
NP grammar is read from a ``treebank'' corpus; then the grammar is
improved by selecting rules with high ``benefit'' scores. Our benefit
measure is identical to that used in transformation-based learning to
select an ordered set of useful transformations \cite{brill:rules-cl}.

Using this simple algorithm with a naive heuristic for matching rules,
we achieve surprising accuracy in an evaluation on two base NP corpora
of varying complexity, both derived from the Penn Treebank WSJ. The
first base NP corpus is that used in the Ramshaw \& Marcus work. The
second espouses a slightly simpler definition of base NP that conforms
to the base NPs used in our Empire sentence analyzer. These simpler
phrases appear to be a good starting point for partial parsers that
purposely delay all complex attachment decisions to later phases of
processing.

Overall results for the approach are promising. For the Empire corpus,
our base NP finder achieves 94\% precision and recall; for the Ramshaw
\& Marcus corpus, it obtains 91\% precision and recall, which is 2\%
less than the best published results. Ramshaw \& Marcus, however,
provide the learning algorithm with word-level information in addition
to the part-of-speech information used in our base NP finder.  By
controlling for this disparity in available knowledge sources, we find
that our base NP algorithm performs comparably, achieving slightly
worse precision (-1.1\%) and slightly better recall (+0.2\%) than the
Ramshaw \& Marcus approach. Moreover, our approach offers many
important advantages that make it appropriate for many NLP tasks:
\begin{itemize}
\item
Training is exceedingly simple.

\item
The base NP bracketer is very fast, operating in time linear in the
length of the text.

\item
The accuracy of the treebank approach is good for applications
that require or prefer fairly simple base NPs.

\item
The learned grammar is easily modified for use with corpora that
differ from the training texts.  Rules can be selectively added to
or deleted from the grammar without worrying about ordering effects.

\item
Finally, our benefit-based training phase offers a simple, general
approach for extracting grammars other than noun phrase grammars from
annotated text.

\end{itemize}

Note also that the treebank approach to base NP identification obtains
good results in spite of a very simple algorithm for ``parsing'' base
NPs.  This is extremely encouraging, and our evaluation suggests at
least two areas for immediate improvement.  First, by replacing the
naive match heuristic with a probabilistic base NP parser that
incorporates lexical preferences, we would expect a nontrivial
increase in recall and precision.  Second, many of the remaining base
NP errors tend to follow simple patterns; these might be corrected
using localized, learnable repair rules.

The remainder of the paper describes the specifics of the approach and
its evaluation. The next section presents the training and application
phases of the treebank approach to base NP identification in more
detail.  Section~\ref{pruning} describes our general approach for
pruning the base NP grammar as well as two instantiations of that
approach.  The evaluation and a discussion of the results appear in
Section~\ref{evaluation}, along with techniques for reducing training
time and an initial investigation into the use of local repair
heuristics.

\begin{figure*}%[tb]
\begin{center}
\epsfxsize= 6.0 in
\hspace*{\fill} 
\epsffile{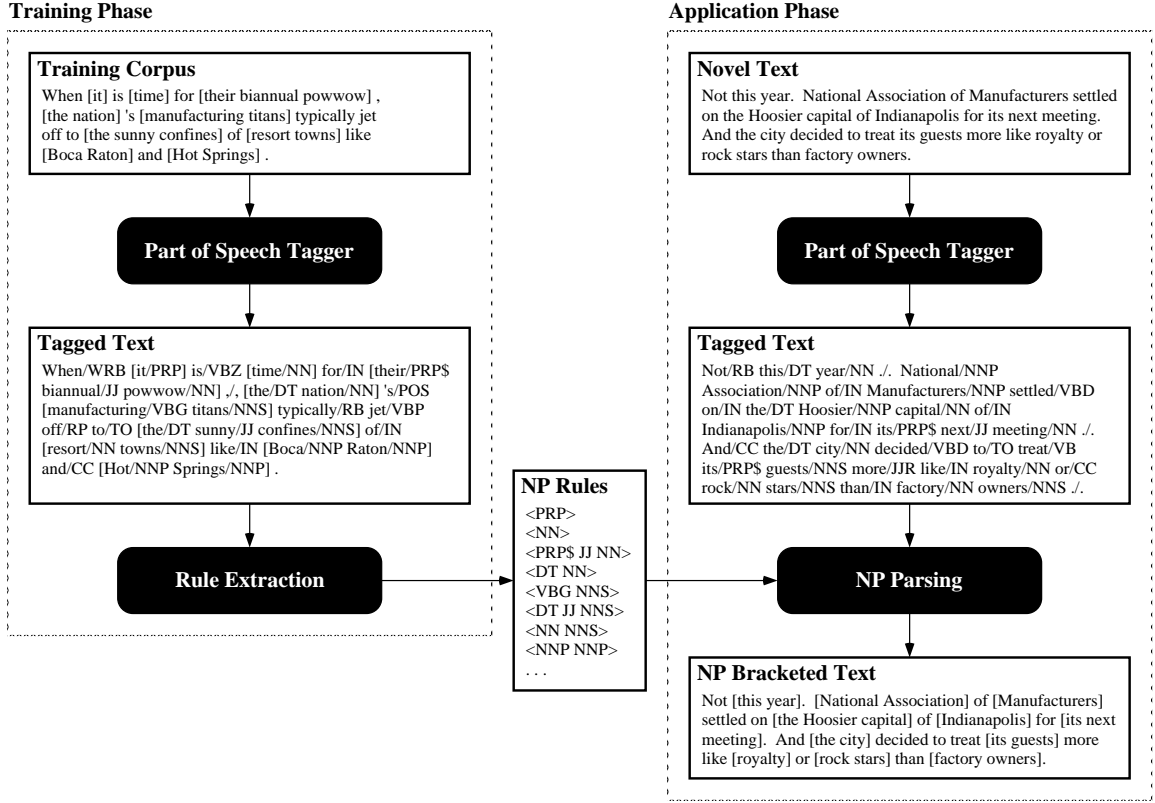}
\hspace*{\fill}
\end{center}
\caption{The Treebank Approach to Base NP Identification}
\label{fig:approach}
\end{figure*}

\section{The Treebank Approach}
\label{approach}

\newcommand{\bij}[2]{b_{({#1}, {#2})}}

Figure~\ref{fig:approach} depicts the treebank approach to base NP
identification. For training, the algorithm requires a corpus that has
been annotated with base NPs.  More specifically, we assume that the
training corpus is a sequence of words $w_1, w_2, \ldots$, along with
a set of base NP annotations $\bij{i_1}{j_1}, \bij{i_2}{j_2}, \ldots$,
where $\bij{i}{j}$ indicates that the NP brackets words $i$ through
$j$: \brnp{$w_i$, \ldots, $w_j$}. The goal of the training phase is to
create a base NP grammar from this training corpus:
\begin{enumerate}
\item Using any available part-of-speech tagger, assign a part-of-speech
tag $t_i$ to each word $w_i$ in the training corpus.

\item Extract from each base noun phrase $\bij{i}{j}$ in the training
corpus its sequence of part-of-speech tags $t_i, \ldots, t_j$ to form
base NP rules, one rule per base NP.

\item Remove any duplicate rules.
\end{enumerate}
The resulting ``grammar'' can then be used to identify base NPs in a novel
text.
\begin{enumerate}
\item
Assign part-of-speech tags $t_1, t_2, \ldots$ to the input words $w_1,
w_2, \ldots$

\item
Proceed through the tagged text from left to right, at each point
matching the NP rules against the remaining part-of-speech tags $t_i,
t_{i+1}, \ldots$ in the text.

\item
If there are multiple rules that match beginning at $t_i$, use the
longest matching rule $R$.  Add the new base noun phrase
$\bij{i}{i+|R|-1}$ to the set of base NPs.  Continue matching at
$t_{i+|R|}$.
\end{enumerate}

With the rules stored in an appropriate data structure, this greedy
``parsing'' of base NPs is very fast. In our implementation, for
example, we store the rules in a decision tree, which permits base NP
identification in time linear in the length of the tagged input text
when using the longest match heuristic.

Unfortunately, there is an obvious problem with the algorithm
described above.  There will be many unhelpful rules in the rule set
extracted from the training corpus.  These ``bad'' rules arise from
four sources: bracketing errors in the corpus; tagging errors; unusual
or irregular linguistic constructs (such as parenthetical
expressions); and inherent ambiguities in the base NPs --- in spite of
their simplicity.  For example, the rule \nprule{VBG NNS}, which was
extracted from \eg{manufacturing/VBG titans/NNS} in the example text,
is ambiguous, and will cause erroneous bracketing in sentences such as
\eg{The execs squeezed in a few meetings before \br{boarding/VBG
buses/NNS} again.}  In order to have a viable mechanism for
identifying base NPs using this algorithm, the grammar must be
improved by removing problematic rules. The next section presents two
such methods for automatically pruning the base NP grammar.

\section{Pruning the Base NP Grammar}
\label{pruning}

\begin{figure}%[tbhp]
\begin{center}
\epsfysize= 2.2in
\hspace*{\fill} 
\epsffile{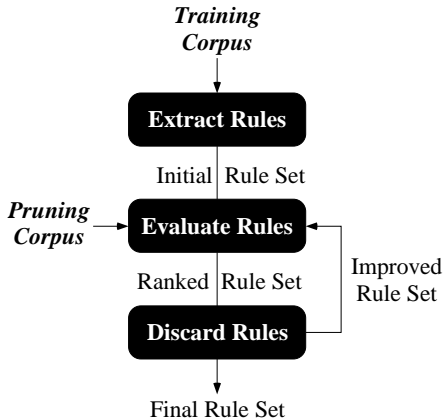}
\hspace*{\fill}
\end{center}
\caption{Pruning the Base NP Grammar}
\label{fig:pruning}
\end{figure}

As described above, our goal is to use the base NP corpus to extract
and select a set of noun phrase rules that can be used to accurately
identify base NPs in novel text.  Our general pruning procedure is
shown in Figure~\ref{fig:pruning}. First, we divide the base NP corpus
into two parts: a training corpus and a pruning corpus. The initial
base NP grammar is extracted from the training corpus as described in
Section~\ref{approach}. Next, the pruning corpus is used to evaluate
the set of rules and produce a ranking of the rules in terms of their
utility in identifying base NPs.  More specifically, we use the rule
set and the longest match heuristic to find all base NPs in the
pruning corpus. Performance of the rule set is measured in terms of
labeled precision ($P$):
\[
P = \frac{\mbox{\# of correct proposed NPs}}{\mbox{\# of proposed NPs}}
\]
We then assign to each rule a score that denotes the ``net benefit''
achieved by using the rule during NP parsing of the improvement
corpus.  The benefit of rule $r$ is given by
$
B_r = C_r - E_r
$
where $C_r$ is the number of NPs correctly identified by $r$, and
$E_r$ is the number of precision errors for which $r$ is
responsible.\footnote{This same benefit measure is also used in the
R\&M study, but it is used to rank transformations rather than to rank
NP rules.}  A rule is considered responsible for an error if it was
the first rule to bracket part of a reference NP, i.e., an NP in the
base NP training corpus.  Thus, rules that form erroneous bracketings
are not penalized if another rule previously bracketed part of the
same reference NP.

For example, suppose the fragment containing base NPs \eg{Boca Raton},
\eg{Hot Springs}, and \eg{Palm Beach} is bracketed as shown below.
\begin{quote}
resort towns like \\ \br{$_{\mathrm{NP}_1}$ Boca/NNP Raton/NNP ,
Hot/NNP} \\ \br{$_{\mathrm{NP}_2}$ Springs/NNP}, and \\
\br{$_{\mathrm{NP}_3}$ Palm/NNP Beach/NNP}
\end{quote}
Rule \nprule{NNP NNP , NNP} brackets NP$_1$; \nprule{NNP} brackets
NP$_2$; and \nprule{NNP NNP} brackets NP$_3$.  Rule \nprule{NNP NNP ,
NNP} incorrectly identifies \eg{Boca Raton , Hot} as a noun phrase, so
its score is $-1$.  Rule \nprule{NNP} incorrectly identifies
\eg{Springs}, but it is not held responsible for the error because of
the previous error by \nprule{NNP NNP , NNP} on the same original NP
\eg{Hot Springs}: so its score is 0.  Finally, rule \nprule{NNP NNP}
receives a score of 1 for correctly identifying \eg{Palm Beach} as a
base NP.

The benefit scores from evaluation on the pruning corpus are used to
rank the rules in the grammar.  With such a ranking, we can improve
the rule set by discarding the worst rules.  Thus far, we have
investigated two iterative approaches for discarding rules, a
\emph{thresholding} approach and an \emph{incremental} approach.  We
describe each, in turn, in the subsections below.

\subsection{Threshold Pruning}

Given a ranking on the rule set, the threshold algorithm simply
discards rules whose score is less than a predefined threshold $R$.
For all of our experiments, we set $R = 1$ to select rules that
propose more correct bracketings than incorrect.  The process of
evaluating, ranking, and discarding rules is repeated until no rules
have a score less than $R$.  For our evaluation on the WSJ corpus,
this typically requires only four to five iterations.

\subsection{Incremental Pruning}

Thresholding provides a very coarse mechanism for pruning the NP
grammar.  In particular, because of interactions between the rules
during bracketing, thresholding discards rules whose score might
increase in the absence of other rules that are also being discarded.
Consider, for example, the \eg{Boca Raton} fragments given earlier. In
the absence of \nprule{NNP NNP , NNP}, the rule \nprule{NNP NNP} would
have received a score of three for correctly identifying all three
NPs.

As a result, we explored a more fine-grained method of discarding
rules: Each iteration of incremental pruning discards the $N$ worst
rules, rather than all rules whose rank is less than some threshold.
In all of our experiments, we set $N = 10$.  As with thresholding, the
process of evaluating, ranking, and discarding rules is repeated, this
time until precision of the current rule set on the pruning corpus
begins to drop.  The rule set that maximized precision becomes the
final rule set.

\subsection{Human Review}

In the experiments below, we compare the thresholding and incremental
methods for pruning the NP grammar to a rule set that was pruned by
hand.  When the training corpus is large, exhaustive review of the
extracted rules is not practical. This is the case for our initial
rule set, culled from the WSJ corpus, which contains approximately
4500 base NP rules.  Rather than identifying and discarding individual
problematic rules, our reviewer identified problematic \emph{classes}
of rules that could be removed from the grammar automatically.  In
particular, the goal of the human reviewer was to discard rules that
introduced ambiguity or corresponded to overly complex base NPs.
Within our partial parsing framework, these NPs are better identified
by more informed components of the NLP system.  Our reviewer
identified the following classes of rules as possibly troublesome:
rules that contain a preposition, period, or colon; rules that contain
WH tags; rules that begin/end with a verb or adverb; rules that
contain pronouns with any other tags; rules that contain misplaced
commas or quotes; rules that end with adjectives.
Rules covered under any of these classes were omitted from the
human-pruned rule sets used in the experiments of
Section~\ref{evaluation}.

\section{Evaluation}
\label{evaluation}

To evaluate the treebank approach to base NP identification, we
created two base NP corpora.  Each is derived from the Penn Treebank
WSJ.  The first corpus attempts to duplicate the base NPs used
the Ramshaw \& Marcus (R\&M) study. The second corpus contains
slightly less complicated base NPs --- base NPs that are better suited
for use with our sentence analyzer, Empire.\footnote{Very briefly, the
Empire sentence analyzer relies on partial parsing to find simple
constituents like base NPs and verb groups. Machine learning
algorithms then operate on the output of the partial parser to perform
all attachment decisions. The ultimate output of the parser is a
semantic case frame representation of the functional structure of the
input sentence.}  By evaluating on both corpora, we can measure the
effect of noun phrase complexity on the treebank approach to base NP
identification. In particular, we hypothesize that the treebank
approach will be most appropriate when the base NPs are sufficiently
simple.

For all experiments, we derived the training, pruning, and testing
sets from the 25 sections of Wall Street Journal distributed with the
Penn Treebank II.  All experiments employ 5-fold cross validation.
More specifically, in each of five runs, a different fold is used for
testing the final, pruned rule set; three of the remaining folds
comprise the training corpus (to create the initial rule set); and the
final partition is the pruning corpus (to prune bad rules from the
initial rule set).  All results are averages across the five folds.
Performance is measured in terms of precision and recall.  Precision
was described earlier --- it is a standard measure of accuracy.
Recall, on the other hand, is an attempt to measure coverage:
\begin{eqnarray*}
P & = & \frac{\mbox{\# of correct proposed NPs}}{\mbox{\# of proposed NPs}}
\\[1ex]
R & = & \frac{\mbox{\# of correct proposed NPs}}{\mbox{\# of NPs in the annotated text}}
\end{eqnarray*}

\begin{table*}
\begin{center}
\begin{small}
\begin{tabular}{|c||c|c|c|c|} \hline
Base NP & Initial     & Threshold   & Incremental & Human \\ 
Corpus  & Rule Set    & Pruning     & Pruning     & Review \\ \hline \hline
Empire  & 23.0P/46.5R & 91.2P/93.1R & 92.7P/93.7R & 90.3P/90.5R \\ \hline    
R\&M    & 19.0P/36.1R & 87.2P/90.0R & 89.4P/90.9R & 81.6P/85.0R \\ \hline
\end{tabular}
\end{small}
\end{center}
\caption{Evaluation of the Treebank Approach Using the Mitre
Part-of-Speech Tagger (P = precision; R = recall)}
\label{tab:results}
\end{table*}

Table~\ref{tab:results} summarizes the performance of the treebank
approach to base NP identification on the R\&M and Empire corpora
using the initial and pruned rule sets.  The first column of results
shows the performance of the initial, unpruned base NP grammar.  The
next two columns show the performance of the automatically pruned rule
sets.  The final column indicates the performance of rule sets that
had been pruned using the handcrafted pruning heuristics.  As
expected, the initial rule set performs quite poorly.  Both automated
approaches provide significant increases in both recall and precision.
In addition, they outperform the rule set pruned using handcrafted
pruning heuristics.

Throughout the table, we see the effects of base NP complexity --- the
base NPs of the R\&M corpus are substantially more difficult for our
approach to identify than the simpler NPs of the Empire corpus.  For
the R\&M corpus, we lag the best published results (93.1P/93.5R) by
approximately 3\%.  This straightforward comparison, however, is not
entirely appropriate.  Ramshaw \& Marcus allow their learning
algorithm to access word-level information in addition to
part-of-speech tags.  The treebank approach, on the other hand, makes
use only of part-of-speech tags.  Table~\ref{tab:without-lexical}
compares Ramshaw \& Marcus' \shortcite{ramshaw:chunking-long} results
with and without lexical knowledge.  The first column reports their
performance when using lexical templates; the second when lexical
templates are not used; the third again shows the treebank approach
using incremental pruning.  The treebank approach and the R\&M
approach without lecial templates are shown to perform comparably
(-1.1P/+0.2R).  Lexicalization of our base NP finder will be addressed
in Section~\ref{repair}.

\begin{table}
\begin{center}
\begin{small}
\begin{tabular}{|c|c|c|} \hline
R\&M (1998)       & R\&M (1998)       & Treebank \\
with              & without           & Approach \\
lexical templates & lexical templates & \\ \hline \hline
93.1P/93.5R       & 90.5P/90.7R       & 89.4P/90.9R \\ \hline
\end{tabular}
\end{small}
\end{center}
\caption{Comparison of Treebank Approach with Ramshaw \& Marcus (1998)
both With and Without Lexical Templates, on the R\&M Corpus}
\label{tab:without-lexical}
\end{table}

Finally, note the relatively small difference between the threshold
and incremental pruning methods in Table~\ref{tab:results}.  For
some applications, this minor drop in performance may be worth the
decrease in training time.  Another effective technique to speed up
training is motivated by Charniak's \shortcite{charniak:treebank}
observation that the benefit of using rules that only occurred once in
training is marginal.  By discarding these rules before pruning, we
reduce the size of the initial grammar --- and the time for
incremental pruning --- by 60\%, with a performance drop of only
-0.3P/-0.1R.

\subsection{Errors and Local Repair Heuristics}
\label{repair}

\begin{table*}
\begin{center}
\begin{small}
\begin{tabular}{|c||c|c||c|c|} \hline
Base NP & Threshold   & Threshold      & Incremental & Incremental \\ 
Corpus  & Improvement & + Local Repair & Improvement & + Local Repair \\ \hline \hline
Empire  & 91.2P/93.1R & 92.8P/93.7R & 92.7P/93.7R & 93.7P/94.0R \\ \hline    
R\&M    & 87.2P/90.0R & 89.2P/90.6R & 89.4P/90.9R & 90.7P/91.1R \\ \hline
\end{tabular}
\end{small}
\end{center}
\caption{Effect of Local Repair Heuristics}
\label{tab:heuristics}
\end{table*}

It is informative to consider the kinds of errors made by the treebank
approach to bracketing.  In particular, the errors may indicate
options for incorporating lexical information into the base NP finder.
Given the increases in performance achieved by Ramshaw \& Marcus by
including word-level cues, we would hope to see similar improvements
by exploiting lexical information in the treebank approach.  For each
corpus we examined the first 100 or so errors and found that certain
linguistic constructs consistently cause trouble.  (In the examples
that follow, the bracketing shown is the error.)
\begin{itemize}
\item
Conjunctions. Conjunctions were a major problem in the R\&M
corpus. For the Empire corpus, conjunctions of adjectives proved
difficult: \eg{\br{record/NN} \br{third-quarter/JJ and/CC
nine-month/JJ results/NNS}}.
\item
Gerunds.  Even though the most difficult VBG constructions such
as \eg{manufacturing titans} were removed from the Empire corpus,
there were others that the bracketer did not handle, like \eg{\br{chief}
operating \br{officer}}. Like conjunctions, gerunds posed a major
difficulty in the R\&M corpus.
\item
NPs Containing Punctuation.  Predictably, the bracketer has difficulty
with NPs containing periods, quotation marks, hyphens, and parentheses.
\item
Adverbial Noun Phrases.  Especially temporal NPs such as
\eg{last month} in \eg{at \br{83.6\%} of \br{capacity last month}}.
\item
Appositives.  These are juxtaposed NPs such as \eg{of \br{colleague
Michael Madden}} that the bracketer mistakes for a single NP.
\item
Quantified NPs.  NPs that look like PPs are a problem: \eg{at/IN
\br{least/JJS} \br{the/DT right/JJ jobs/NNS}}; \eg{about/IN \br{25/CD
million/CD}}.
\end{itemize}

Many errors appear to stem from four underlying causes.  First, close
to 20\% can be attributed to errors in the Treebank and in the Base NP
corpus, bringing the effective performance of the algorithm to
94.2P/95.9R and 91.5P/92.7R for the Empire and R\&M corpora,
respectively.  For example, neither corpus includes WH-phrases as base
NPs.  When the bracketer correctly recognizes these NPs, they are
counted as errors.  Part-of-speech tagging errors are a second cause.
Third, many NPs are missed by the bracketer because it lacks the
appropriate rule.  For example, \eg{household products business} is
bracketed as \eg{\br{household/NN products/NNS}
\br{business/NN}}. Fourth, idiomatic and specialized expressions,
especially time, date, money, and numeric phrases, also account for a
substantial portion of the errors.

These last two categories of errors can often be detected because
they produce either recognizable patterns or unlikely linguistic
constructs.  Consecutive NPs, for example, usually denote bracketing
errors, as in \eg{\br{household/NN products/NNS}
\br{business/NN}}. Merging consecutive NPs in the correct contexts
would fix many such errors.  Idiomatic and specialized expressions
might be corrected by similarly local repair heuristics.  Typical
examples might include changing \eg{\br{effective/JJ Monday/NNP}} to
\eg{effective \br{Monday}}; changing \eg{\br{the/DT balance/NN
due/JJ}} to \eg{\br{the balance} due}; and changing \eg{were/VBP
\br{n't/RB the/DT only/RB losers/NNS}} to \eg{were n't \br{the only
losers}}.

Given these observations, we implemented three local repair
heuristics.  The first merges consecutive NPs unless either might be a
time expression.  The second identifies two simple date expressions.
The third looks for quantifiers preceding \eg{of NP}. The first
heuristic, for example, merges \eg{\br{household products}
\br{business}} to form \eg{\br{household products business}}, but
leaves \eg{increased \br{15 \%} \br{last Friday}} untouched.  The
second heuristic merges \eg{\br{June 5} , \br{1995}} into \eg{\br{June
5, 1995}}; and \eg{\br{June} , \br{1995}} into \eg{\br{June, 1995}}.
The third finds examples like \eg{some of \br{the companies}} and
produces \eg{\br{some} of \br{the companies}}.  These heuristics
represent an initial exploration into the effectiveness of employing
lexical information in a post-processing phase rather than during
grammar induction and bracketing.  While we are investigating the
latter in current work, local repair heuristics have the advantage of
keeping the training and bracketing algorithms both simple and fast.

The effect of these heuristics on recall and precision is shown in
Table~\ref{tab:heuristics}.  We see consistent improvements for both
corpora and both pruning methods, achieving approximately 94P/R for
the Empire corpus and approximately 91P/R for the R\&M corpus.  Note
that these are the final results reported in the introduction and
conclusion.  Although these experiments represent only an initial
investigation into the usefulness of local repair heuristics, we are
very encouraged by the results.  The heuristics uniformly boost
precision without harming recall; they help the R\&M corpus even
though they were designed in response to errors in the Empire corpus.
In addition, these three heuristics alone recover 1/2 to 1/3 of the
improvements we can expect to obtain from lexicalization based on the
R\&M results.

\section{Conclusions}
\label{conclusions}

This paper presented a new method for identifying base NPs.  Our
treebank approach uses the simple technique of matching part-of-speech
tag sequences, with the intention of capturing the simplicity of the
corresponding syntactic structure.  It employs two existing
corpus-based techniques: the initial noun phrase grammar is extracted
directly from an annotated corpus; and a benefit score calculated from
errors on an improvement corpus selects the best subset of rules via a
coarse- or fine-grained pruning algorithm.

The overall results are surprisingly good, especially considering the
simplicity of the method. It achieves 94\% precision and recall on
simple base NPs. It achieves 91\% precision and recall on the more
complex NPs of the Ramshaw \& Marcus corpus. We believe, however, that
the base NP finder can be improved further.
First, the longest-match heuristic of the noun phrase bracketer could
be replaced by more sophisticated parsing methods that account for
lexical preferences.  Rule application, for example, could be
disambiguated statistically using distributions induced during
training.  We are currently investigating such extensions.  One
approach closely related to ours --- weighted finite-state transducers
(e.g.\ \cite{pereira:wfst:1997}) --- might provide a principled way to
do this.  We could then consider applying our error-driven pruning
strategy to rules encoded as transducers.
Second, we have only recently begun to explore the use of local repair
heuristics. While initial results are promising, the full impact of
such heuristics on overall performance can be determined only if they
are systematically learned and tested using available training data.
Future work will concentrate on the corpus-based acquisition of local
repair heuristics.

In conclusion, the treebank approach to base NPs provides an accurate
and fast bracketing method, running in time linear in the length of
the tagged text.  The approach is simple to understand, implement, and
train. The learned grammar is easily modified for use with new
corpora, as rules can be added or deleted with minimal interaction
problems. Finally, the approach provides a general framework for
developing other treebank grammars (e.g., for subject/verb/object
identification) in addition to these for base NPs.

\begin{small}
\paragraph{Acknowledgments.} This work was supported in part by NSF
Grants IRI--9624639 and GER--9454149.  We thank Mitre for providing
their part-of-speech tagger.
\end{small}

\bibliographystyle{acl}

\end{document}